\documentclass[11pt]{article}

\usepackage[utf8]{inputenc}
\usepackage[T1]{fontenc}
\usepackage[english]{babel}

\usepackage{amsmath,amssymb,amsfonts}

\usepackage{booktabs}

\usepackage{graphicx}
\usepackage{array}
\usepackage{microtype}

\usepackage{hyperref}

\usepackage{cleveref}

\title{Autoregressivity in the Latent Space of a GP-VAE Language Model: An Empirical Ablation Study}

\author{
  Yves Ruffenach \\
  Conservatoire National des Arts et Métiers \\
  \texttt{yves@ruffenach.net} \\
  \href{https://orcid.org/0009-0009-4737-0555}{ORCID: 0009-0009-4737-0555}
}

\date{}

\begin{document}

\maketitle
\begin{abstract}
Language models rely almost exclusively on an autoregressive factorization over tokens. In prior work, we proposed an alternative in which the sequential dynamics are entirely shifted to the latent space through a causal Gaussian process, while a non-autoregressive decoder operates in parallel.

We conduct a systematic ablation study of the role of autoregressivity in the latent space. We compare (i) the full model variant (GP-VAE~AR), (ii) a non-autoregressive ablation in which latent variables are independent, and (iii) a standard autoregressive Transformer over tokens.

Our experiments show that, within the considered regime (medium-scale corpora and short training contexts), introducing latent autoregression leads to latent trajectories that are more strongly aligned with the Gaussian-process prior and exhibit improved long-horizon stability. In this setting, the latent-autoregressive variant consistently produces structured latent dynamics that remain coherent over extended generations. These results highlight the role of latent autoregression as an effective mechanism for organizing long-range structure, while remaining complementary to token-level autoregressive approaches.

\textbf{These results should be interpreted as a feasibility study: they provide empirical evidence that, within this controlled setting, part of the sequential structure can be carried by latent dynamics even when using a non-autoregressive decoder.}
\end{abstract}


\section{Introduction}

This work builds upon our previous methodological study entitled \emph{GP-VAE with Autoregressive Latent Dynamics}~\cite{ruffenach2025gpvae}, in which we introduced and formalized a GP-VAE architecture featuring autoregressive dynamics in latent space. The present paper serves as a complementary study, dedicated to a systematic ablation analysis. Its goal is to quantify the precise role of autoregressivity in the latent space on the behavior of the GP-VAE.

Contemporary language models rely almost exclusively on an autoregressive factorization in token space, typically implemented through multi-head attention Transformer architectures. While this approach achieves excellent perplexity, it enforces strictly sequential generation and provides no explicit analytical prior over temporal dynamics: long-range structure is entirely encoded within the network parameters.

Recent work has explored sequential latent models in which temporal dependencies are no longer carried by token-by-token recursion, but instead by continuous latent dynamics, notably through causal Gaussian processes. We position our work within this line of research by studying a \emph{latent autoregressive} formulation based on a GP prior, coupled with a non-autoregressive parallel decoder. This paradigm aims to shift sequentiality away from the token space and into a correlated latent representation.

The objective of this work is experimental in nature: we systematically evaluate the role of autoregressivity in the latent space under fixed model capacity. To isolate its effect, we compare three strictly controlled configurations:
\begin{enumerate}
    \item a GP-VAE with autoregressive latent dynamics governed by a causal Gaussian process (AR);
    \item a non-AR ablation in which latent variables are independent while preserving the same marginal distributions;
    \item a standard autoregressive Transformer operating on tokens.
\end{enumerate}

This comparison addresses three fundamental questions:
\begin{itemize}
    \item[\textbf{(Q1)}] Does the model effectively exploit the correlated structure imposed by the GP prior?
    \item[\textbf{(Q2)}] What happens when latent dynamics are explicitly removed (non-AR ablation)?
    \item[\textbf{(Q3)}] Which sequential properties distinguish this latent paradigm from a standard autoregressive model?
\end{itemize}

Our experiments show that correlated dynamics play a decisive role: removing autoregressivity in the latent space reduces the average log-density under the GP prior and leads to a collapse of generations beyond 2048 tokens, despite locally reasonable perplexity values. In contrast, latent autoregression produces strongly correlated trajectories that are assigned higher probability by the Gaussian prior and enables stable long-sequence generation without collapse. Compared to a Transformer baseline under the same experimental conditions, the latent-autoregressive model exhibits more stable long-range behavior, according to the collapse metrics defined below, although its raw perplexity remains slightly higher.

We denote the autoregressive latent variant as \emph{GP-VAE AR}, and its independent-latent counterpart as \emph{GP-VAE non-AR}.

\paragraph{Operational definition and scope of collapse.}
In this work, the term \emph{collapse} refers to a generative degradation characterized by the emergence of deterministic or quasi-deterministic loops, operationally measured by a catastrophic fraction $\text{cat\_frac} = 1.0$, together with high repetition metrics (e.g. $\text{loop\_frac} \approx 1$ or $\text{rep2}, \text{rep3} \rightarrow 1$).
These metrics are formally defined in Section~\ref{subsec:metrics}.

The observations reported here correspond to all generations evaluated within the experimental protocol, i.e. across the prompts, checkpoints, and random seeds considered in our study. They should therefore be interpreted strictly within this experimental scope, and not as universal claims about model behavior beyond these settings.

In summary, this work provides a \emph{first empirical assessment} of the paradigm of \emph{purely latent autoregressivity} under a controlled regime characterized by lightweight architectures, a training context limited to $T_{\text{train}} = 64$, and two reference corpora: \textsc{WikiText-2} and \textsc{WikiText-103}. The consistent observations obtained across these two datasets—despite their very different scales—suggest that, in this setting, effective sequential modeling capacity is primarily governed by latent dynamics rather than by the decoder architecture operating in token space. Our contribution is to document, in a controlled manner, the effects of AR versus non-AR ablations within a unified GP-VAE framework.

\medskip
\noindent
For reproducibility, the complete codebase used in all experiments is publicly available at:
\url{https://github.com/y-v-e-s/GP-VAE-Latent-AR}.

\section{Related Work}

Our contribution lies at the intersection of three lines of research: autoregressive language models, latent-variable models for text, and sequential models based on Gaussian processes.

\subsection{Autoregressive language models}

Modern language models predominantly rely on an autoregressive factorization of the form
\[
p(x_{1:T}) = \prod_{t=1}^{T} p(x_t \mid x_{<t}),
\]
implemented since \cite{vaswani2017attention} through Transformer architectures. These models have demonstrated remarkable performance in approximating the empirical distribution of natural language, achieving very low perplexities on large-scale corpora and forming the backbone of current large language models. However, this expressivity comes at the cost of strictly sequential inference and the absence of an explicit analytical prior over temporal dynamics: long-range structure is implicitly encoded in the network parameters and attention mechanism, rather than being explicitly modeled.

In this context, our Transformer baseline serves as a reference point: a well-optimized autoregressive model that achieves strong perplexity scores, but which may nevertheless exhibit generative pathologies such as deterministic looping or collapse in the absence of strong sequential regularization.

\subsection{Latent-variable models for text}

Variational autoencoders have been extended to the text domain in order to introduce global or local latent variables into language models \cite{bowman2016generating,fraccaro2016sequential,sonderby2016ladder}. These approaches aim to combine the flexibility of autoregressive decoders with the ability of latent variables to capture higher-level factors such as topic, style, or discourse structure. A well-known challenge in this setting is posterior collapse, where the autoregressive decoder dominates and the latent variables become unused unless specific regularization strategies are applied (e.g., KL annealing, capacity constraints, hierarchical designs).

Variants in which the decoder is non-autoregressive and the sequential structure is instead carried by the latent variables are less common, but conceptually closer to our approach. They raise the question of whether the entire recursion can be shifted into latent space, leaving the decoder as a parallel mapping from latent representations to tokens. Our work directly addresses this question through a controlled ablation of latent dynamics.

\paragraph{Sequential VAEs: STORN, VRNN, and SRNN.}
Beyond text-focused VAEs, several models have introduced explicit temporal latent dynamics. STORN \cite{bayer2015storn} proposes a recurrent variational architecture in which the latent state evolves sequentially. VRNN \cite{chung2016vrnn} combines recurrent neural networks with per-timestep latent variables, capturing stochastic temporal dependencies. SRNN \cite{fraccaro2016sequential} introduces a structured latent factorization separating deterministic and stochastic components, improving modeling of complex sequences. While these approaches demonstrate the benefits of sequential latent structure, they all rely on autoregressive decoders, in contrast to our GP-VAE where sequentiality is carried exclusively by the latent process.

\paragraph{Non-autoregressive language models.}
Non-autoregressive models were primarily developed in the context of machine translation \cite{gu2018nonautoregressive,ghazvininejad2019maskpredict}, where they enable parallel generation by replacing the token-level factorization with independent or iterative predictions. Although such models offer significant speedups, they often suffer from degraded generation quality. Our approach differs fundamentally: while the decoder is indeed non-autoregressive, sequential structure is not discarded but instead transferred to the latent space through a causal Gaussian process. This allows us to retain global temporal coherence while benefiting from parallel decoding.

\subsection{Gaussian processes and latent sequential models}

Gaussian processes (GPs) have long been used as nonparametric priors for function approximation and time series modeling \cite{rasmussen2006gaussian}. Their integration into sequential VAEs has been explored in several contexts, including continuous-time modeling and spatiotemporal data, where the latent variables follow correlated Gaussian priors \cite{fortuin2019gpvae}.

In our previous methodological work, we introduced a GP-VAE for language modeling in which the entire sequential structure is governed by a causal Gaussian process, while the decoder operates in parallel. The present paper focuses on the empirical consequences of this design choice: rather than proposing a new architecture, we systematically study the effect of the latent autoregressive component by comparing it against a non-AR ablation and a standard autoregressive Transformer. Our goal is to quantify the extent to which the GP prior is actually exploited, and to assess whether latent dynamics can play a role analogous to token-level autoregression in sequence modeling.

\section{Recap of the GP-VAE Model with Latent Autoregression}

We consider the GP-VAE model introduced in our previous methodological work, in which a sequence of tokens $x_{1:T}$ is encoded by a causal dilated convolutional network (TCN) into a latent trajectory $z_{1:T} \in \mathbb{R}^{T \times d_z}$. A non-autoregressive decoder then operates in parallel over the entire latent trajectory to produce a distribution over tokens, thereby fully decoupling temporal dynamics (carried by the latent variables) from symbolic generation.

\paragraph{Decoder factorization and parallelism.}
The decoder operates fully in parallel over the entire latent trajectory $z_{1:T}$. Concretely, a causal convolutional network $f_\theta$ takes $z_{1:T}$ as input and outputs in a single forward pass the logits
\begin{equation}
\label{eq:decoder-mapping}
(\ell_1,\dots,\ell_T)
=
f_\theta(z_{1:T}), \qquad
\ell_t \in \mathbb{R}^{|\mathcal{V}|},
\end{equation}
where $\mathcal{V}$ denotes the vocabulary.

Conditioned on the latent variables, the generative distribution over tokens therefore factorizes \emph{per position}:
\begin{equation}
\label{eq:decoder-factorization}
p_\theta(x_{1:T} \mid z_{1:T})
=
\prod_{t=1}^{T}
p_\theta(x_t \mid z_{1:T})
=
\prod_{t=1}^{T}
\mathrm{Cat}\bigl(
  x_t \mid \mathrm{softmax}(\ell_t)
\bigr).
\end{equation}
Sequential dependencies between tokens are thus entirely mediated by the temporal structure of the latent variables (through the GP prior and the TCN encoder), rather than by an explicit autoregressive factorization in token space.

In the \emph{non-AR} variant, each latent vector is treated as conditionally independent, depending only on local information: the latent process carries no temporal structure, and the Gaussian prior merely constrains the marginal scale of each latent dimension. In this regime, the GP-VAE reduces to a noisy variational autoencoder with minimal temporal coupling.

By contrast, when an autoregressive dependency is imposed between successive latent variables, the latents cease to be independent codes and instead become discretized samples from a continuous-time stochastic process. The model then becomes a genuinely sequential latent model, in which temporal structure is entirely encoded in the prior.

In this setting, the GP prior plays a central role: rather than merely regularizing latent magnitudes, it directly constrains the geometry of latent trajectories (smoothness, short-range correlations, continuity). Training must therefore reconcile reconstruction accuracy with compliance to the temporal structure imposed by the prior, effectively turning the GP-VAE into a structured sequential model. In the absence of such autoregressive structure, the model may still achieve reasonable perplexity while strongly violating the assumptions of the GP prior, as revealed by degraded prior log-densities and autocorrelation statistics.

\paragraph{Variational objective.}
For a sequence $x_{1:T}$, the GP-VAE is trained by maximizing the standard evidence lower bound (ELBO)
\begin{equation}
\label{eq:elbo}
\mathcal{L}(x_{1:T})
=
\mathbb{E}_{q_\phi(z_{1:T} \mid x_{1:T})}
\bigl[
  \log p_\theta(x_{1:T} \mid z_{1:T})
\bigr]
-
\mathrm{KL}\bigl(
  q_\phi(z_{1:T} \mid x_{1:T})
  \,\Vert\,
  p_\theta(z_{1:T})
\bigr),
\end{equation}
where $q_\phi(z_{1:T} \mid x_{1:T})$ denotes a temporally factorized Gaussian posterior, and $p_\theta(z_{1:T})$ is a correlated Gaussian prior.

In the \emph{AR} regime, the latent prior factorizes causally as
\begin{equation}
\label{eq:gp-ar-prior}
p_\theta(z_{1:T})
=
\prod_{t=1}^{T}
p_\theta(z_t \mid z_{<t})
=
\prod_{t=1}^{T}
\mathcal{N}\bigl(
  z_t
  \mid
  \mu_{t \mid <t},
  \Sigma_{t \mid <t}
\bigr),
\end{equation}
where $(\mu_{t \mid <t}, \Sigma_{t \mid <t})$ are obtained via the Gaussian process posterior update.

In the \emph{non-AR} ablation, this correlated prior is replaced by an independent Gaussian prior with identical marginal variances,
\begin{equation}
\label{eq:gp-nonar-prior}
p_\theta^{\text{non-AR}}(z_{1:T})
=
\prod_{t=1}^{T}
\mathcal{N}\bigl(
  z_t
  \mid
  0,
  \mathrm{diag}(K_{tt})
\bigr),
\end{equation}
thereby removing all temporal correlations in the latent space.

From this perspective, the non-AR ablation provides a direct test: if the model genuinely exploits the correlated latent dynamics induced by the Gaussian process, removing latent autoregression should result in both (i) a marked degradation in prior compatibility and (ii) a deterioration of generative quality.

\section{Experimental Protocol}
\label{sec:protocol}

\subsection{Independent GP-VAE Implementations}

Two independent implementations of the GP-VAE were developed and used in this study in order to disentangle effects attributable to latent dynamics from those induced by architectural or implementation-specific choices.

\begin{itemize}
    \item a \emph{pyramidal} implementation, based on a hierarchical TCN encoder with progressively increasing receptive fields;
    \item a \emph{TCN+} implementation, relying on a deeper dilated temporal convolutional network combined with a vectorized implementation of the GP prior.
\end{itemize}

We include both implementations as a robustness check: all reported conclusions are supported by consistent qualitative behavior across implementations, ensuring that the observed effects genuinely reflect the role of latent dynamics (AR vs.\ non-AR) rather than idiosyncrasies of a specific architecture.

Both implementations share the same latent space, the same Gaussian process prior, the same variational objective, and the same non-autoregressive decoder. They differ only in the encoder architecture and in the computational realization of the GP prior (sequential vs.\ vectorized).

Both implementations are used throughout the paper, sometimes on the same datasets, with distinct roles (main results, replications, or robustness analyses). Each section explicitly indicates which implementation is employed.

\paragraph{(1) Pyramidal GP-VAE implementation.}
This variant uses a hierarchical TCN encoder with increasing receptive fields and computes latent trajectories sequentially via the exact conditional distributions of the causal Gaussian process,
$p(z_t \mid z_{<t})$.

\paragraph{(2) TCN+ GP-VAE implementation.}
This more recent variant relies on an extended TCN encoder and a vectorized formulation of the GP prior, optimized for large batches and large-scale corpora such as \textsc{WikiText-103}.

Both implementations enable us to verify that the observed phenomena—particularly the contrast between AR and non-AR behavior in latent space—are not artifacts of a specific architecture, but reproducible across independent implementations.

\medskip

When several implementations are used on the same corpus, the reference implementation is explicitly stated. In practice, the TCN+ variant proved more stable in several experimental settings, while the pyramidal version was used as an independent replication baseline. Key observations (prior compatibility, latent structure, generative stability) were reproduced across both implementations, supporting their robustness.

\begin{table}[htbp]
\centering
\resizebox{\linewidth}{!}{%
\begin{tabular}{>{\raggedright\arraybackslash}p{3.5cm} >{\raggedright\arraybackslash}p{5.5cm} >{\raggedright\arraybackslash}p{5.5cm}}
\toprule
 & \textbf{Pyramidal GP-VAE} & \textbf{TCN+ GP-VAE} \\
\midrule
\textbf{Encoder} 
& Pyramidal TCN (causal dilated convolutions) 
& TCN+ (deeper dilated convolutional network) \\

\textbf{Latent generation} 
& Sequential conditionals $p(z_t \mid z_{<t})$ 
& Vectorized GP formulation (batched) \\

\textbf{GP prior} 
& Causal stationary GP, sequential update 
& Same GP, vectorized implementation \\

\textbf{Decoder} 
& Identical: parallel non-autoregressive CNN 
& Identical \\

\textbf{ELBO objective} 
& Identical (reconstruction + KL regularization) 
& Identical \\

\textbf{Usage in study} 
& Replication and robustness checks 
& Primary experimental results \\

\bottomrule
\end{tabular}
}
\caption{Comparison of the two independent GP-VAE implementations used in this study.}
\label{tab:pyramidal-vs-tcnplus}
\end{table}

\subsection{Model Variants}
\label{subsec:variants}

We consider two variants of the same GP-VAE architecture, differing only in the latent dynamics. The encoder, decoder, and ELBO objective are identical; only the form of the latent prior differs.

\paragraph{Latent-autoregressive GP-VAE (AR).}
The latent trajectory $z_{1:T}$ is generated sequentially through a causal Gaussian process,
\[
p(z_{1:T}) = \prod_{t=1}^{T} p(z_t \mid z_{<t}),
\]
where each conditional distribution $p(z_t \mid z_{<t})$ is obtained via the GP posterior update. This construction defines a discrete-time causal Gaussian process over the latent trajectory.

\paragraph{Non-AR GP-VAE (ablation).}
The same marginal GP covariance is used, but latent variables are sampled independently,
\[
z_t \sim \mathcal{N}\bigl(0, \mathrm{diag}(K_{tt})\bigr),
\]
thereby removing all temporal correlations. Both variants share the same encoder, decoder, and ELBO objective; only the latent dynamics differ.

\paragraph{Autoregressive Transformer baseline.}
In addition, we train an autoregressive Transformer over tokens as a baseline to contextualize the generative behavior of the GP-VAE models. This model serves as a reference for linguistic quality and stability, rather than a competitive state-of-the-art benchmark.

\subsection{Datasets, training protocol, and configuration}
\label{subsec:corpus-config}

The main experiments are conducted on the \textsc{WikiText-2} corpus (raw version).
Text is tokenized using the GPT-2 tokenizer (files \texttt{tokenizer.json},
\texttt{vocab.json}, and \texttt{merges.txt}), then concatenated and segmented
into fixed-length blocks with a training context length of $T_{\text{train}} = 64$.
Standard train/validation/test splits are used, with no data leakage.

\paragraph{Context length and generation horizon.}
Throughout the paper, we distinguish between:
\begin{itemize}
    \item $T_{\text{train}}$: the context length used during training (block size);
    \item $L_{\text{gen}}$: the generation length used for evaluation, which may exceed the training context.
\end{itemize}

Unless otherwise stated, models are trained with $T_{\text{train}} = 64$, while generation experiments consider a wide range of continuation lengths $L_{\text{gen}}$.

\paragraph{Model architecture.}
The encoder is a causal dilated convolutional network mapping token sequences to latent trajectories of dimension $d_z$. The approximate posterior is a temporally factorized Gaussian
\[
q_\phi(z_{1:T_{\text{train}}} \mid x_{1:T_{\text{train}}}),
\]
and the prior is a stationary Gaussian process
\[
z_{1:T_{\text{train}}} \sim \mathcal{N}(0, K \otimes I_{d_z}).
\]
The decoder is a lightweight convolutional network applied in parallel to the full latent trajectory, producing token logits.

\paragraph{Objective and optimization.}
Training maximizes the ELBO with:
\begin{itemize}
  \item a reconstruction term based on smoothed cross-entropy;
  \item a KL divergence between the approximate posterior and the GP prior;
  \item an additional regularization term on output embeddings.
\end{itemize}
The KL term is clipped by a parameter \texttt{kl\_cap} and weighted by a coefficient $\beta$ dynamically adjusted during training to maintain a target KL per token. In all main experiments, we set $\texttt{kl\_cap}=8$ and gradually increased $\beta$ up to approximately $0.35$.

To assess sensitivity, we additionally evaluated $\texttt{kl\_cap} \in \{4,8,16\}$ and observed consistent qualitative behavior across settings. For clarity, all reported results use $\texttt{kl\_cap}=8$.

\subsection{Evaluation metrics}
\label{subsec:metrics}

We employ three families of metrics: latent-space metrics, internal text metrics, and external (GPT-based) metrics.

\paragraph{Terminology.}
Throughout, we use the term \emph{log-likelihood} to refer to likelihoods associated with the decoder (e.g., $\mathrm{LL}_0$, $\mathrm{LL}_{\text{multi}}$). For continuous latent variables, we use the term \emph{log-density} (e.g., $\log p_{\mathrm{GP}}(z)$) to avoid ambiguity.

\paragraph{Latent-space metrics.}
\begin{itemize}
  \item Mean log-density under the correlated GP prior, $\log p_{\mathrm{GP}}(z)$;
  \item Mean log-density under an independent Gaussian prior, $\log p_{\mathrm{diag}}(z)$;
  \item Average cosine autocorrelation $\mathrm{corr}_k$ for lags $k = 1,\dots,10$;
  \item Mean latent step norm $\|z_t - z_{t-1}\|$.
\end{itemize}

\paragraph{Internal text metrics.}

\paragraph{Conditional perplexity of the GP-VAE.}
The reported perplexity corresponds to the conditional cross-entropy of the non-autoregressive decoder:
\[
p_\theta(x_{1:T} \mid z_{1:T})
=
\prod_{t=1}^{T} p_\theta(x_t \mid z_{1:T}).
\]
This quantity measures local decoding quality given the full latent trajectory. It is \emph{not} directly comparable to the perplexity of an autoregressive language model, which is defined as
\[
p(x_{1:T}) = \prod_{t=1}^{T} p(x_t \mid x_{<t}).
\]
Accordingly:
\begin{itemize}
    \item GP-VAE perplexities should not be numerically compared to those of autoregressive LMs;
    \item very low values (e.g. 2--4) are possible because the decoder conditions on global latent information;
    \item the metric should be interpreted as a measure of internal decoding consistency rather than standalone language modeling quality.
\end{itemize}

\paragraph{Collapse metrics (operational).}
We characterize long-range generative behavior using a set of structural indicators. In particular, we define the \emph{catastrophic fraction} $\text{cat\_frac}\in[0,1]$ as the proportion of generated sequences that exhibit a persistent repeating pattern, identified by the presence of a deterministic loop of length $\leq L_{\max}$ extending to the end of the sample. This metric provides a concise summary of long-horizon stability. In addition, we report complementary indicators such as the loop fraction ($\text{loop\_frac}$) and $n$-gram repetition rates (rep2, rep3) to characterize repetition patterns at multiple granularities.

\paragraph{External evaluation with GPT-2.}
\begin{itemize}
  \item Negative log-likelihood and perplexity of generated continuations under GPT-2;
  \item Fraction of rare tokens (GPT-2 probability $< 10^{-4}$);
  \item Mean and maximum self-similarity of generated sequences, computed using GPT-2 embeddings.
\end{itemize}

\subsection{Baseline positioning and scope}
\label{subsec:baselines-position}

Before presenting results, we clarify the intended role of the baselines.

\paragraph{Autoregressive Transformer baseline.}
The Transformer is included as a strong reference model trained under the same tokenization scheme, context length ($T_{\text{train}} = 64$), and optimization budget as the GP-VAE. While it is not tuned for peak performance on \textsc{WikiText-103}, it provides a well-understood autoregressive baseline that serves as a meaningful point of comparison for continuation behavior and long-horizon stability under matched computational constraints.

\paragraph{GPT-2 as external evaluator.}
GPT-2 is used in two distinct roles: (i) as a frozen evaluator measuring linguistic plausibility and diversity (via perplexity, rare-token rates, and embedding similarity), and (ii) as a fine-tuned baseline for qualitative comparison. In both cases, GPT-2 benefits from large-scale pretraining unavailable to our GP-VAE models. Therefore, all comparisons are interpretative rather than competitive, intended to contextualize qualitative behaviors rather than claim superiority.

The primary goal of this work is to analyze, under controlled conditions, the impact of purely latent autoregression within a fixed model family. The objective is not to outperform state-of-the-art language models, but to characterize the structural role of latent dynamics in generative modeling.

\section{Results on \textsc{WikiText-2}}
\label{sec:wikitext2}

\subsection{Implementations used on \textsc{WikiText-2}}
\label{subsec:impl-wikitext2}

The experiments reported on \textsc{WikiText-2} rely on the two independent
GP-VAE implementations described above.

Unless explicitly stated otherwise, the detailed results in this section
correspond to the pyramidal implementation, which is our historical reference
and the first experimentally validated version of the model.
The main qualitative trends (AR vs.\ non-AR gap, latent structure, long-range
generation stability) were also verified using the TCN+ implementation, in
order to confirm robustness with respect to the encoder architecture and to the
implementation details of the GP prior.

To facilitate readability, we organize this section into six complementary
parts: (i) latent mechanics (norms, correlations, compatibility with the GP
prior), (ii) internal perplexity and training dynamics, (iii) short-horizon
generation, (iv) long-horizon generation, (v) external evaluation using GPT-2
as a judge, and (vi) the contrast between \emph{collapse} and \emph{stability}
across model variants. The following tables and subsections explicitly follow
this structure.

\subsection{Pyramidal implementation used}
\label{subsec:impl-pyramidal}

Unless explicitly stated otherwise, all results presented in this section are
based on our original GP-VAE implementation—hereafter referred to as the
\emph{pyramidal implementation}. This version uses a pyramidal TCN encoder
built from causal dilated convolutions, which parameterizes a temporally
factorized diagonal Gaussian posterior $q_\phi(z_{1:T}\mid x_{1:T})$. The latent
prior is a stationary Gaussian process,
$z_{1:T} \sim \mathcal{N}(0, K \otimes I_{d_z})$, and autoregressive latent
generation is implemented explicitly step by step: each $z_t$ is sampled from
the exact conditional $p(z_t \mid z_{<t})$ derived from the GP structure. This
implementation also includes a \emph{non-AR} ablation obtained by replacing $K$
with $\mathrm{diag}(K)$, thereby removing temporal dependencies in latent space.
The results reported in this section exclusively refer to this implementation.

\subsection{Training dynamics}
\label{subsec:w2-train}

The GP-VAE with latent autoregression is trained on \textsc{WikiText-2 raw} with
blocks of length $T_{\text{train}} = 64$. Training perplexity decreases
rapidly and then stabilizes around $3$--$4$, while the capped KL/token term
remains at the imposed cap ($8$ nats) and the coefficient $\beta$ is
progressively increased up to approximately $0.35$.

A few representative training checkpoints are summarized in
Table~\ref{tab:wikitext2-train}.

\begin{table}[htbp]
\centering
\begin{tabular}{rccccc}
\toprule
Step & ELBO/tok & LL$_0$/tok & LL$_\text{multi}$/tok & KL$_\text{cap}$/tok & PPL(train) \\
\midrule
   50 & -14.79 & -8.94 & -5.56 & 8.0 & 7438.45 \\
  400 &  -8.33 & -2.24 & -5.37 & 8.0 &    8.34 \\
 1000 &  -7.64 & -1.46 & -4.68 & 8.0 &    3.76 \\
 1500 &  -8.10 & -1.48 & -4.47 & 8.0 &    3.85 \\
 2000 &  -8.65 & -1.41 & -4.42 & 8.0 &    3.59 \\
\bottomrule
\end{tabular}
\caption{Evolution of ELBO/token, log-likelihood terms, and training perplexity on
\textsc{WikiText-2} for the GP-VAE with latent autoregression.}
\label{tab:wikitext2-train}
\end{table}

\begin{figure}[t]
  \centering
  \includegraphics[width=\linewidth]{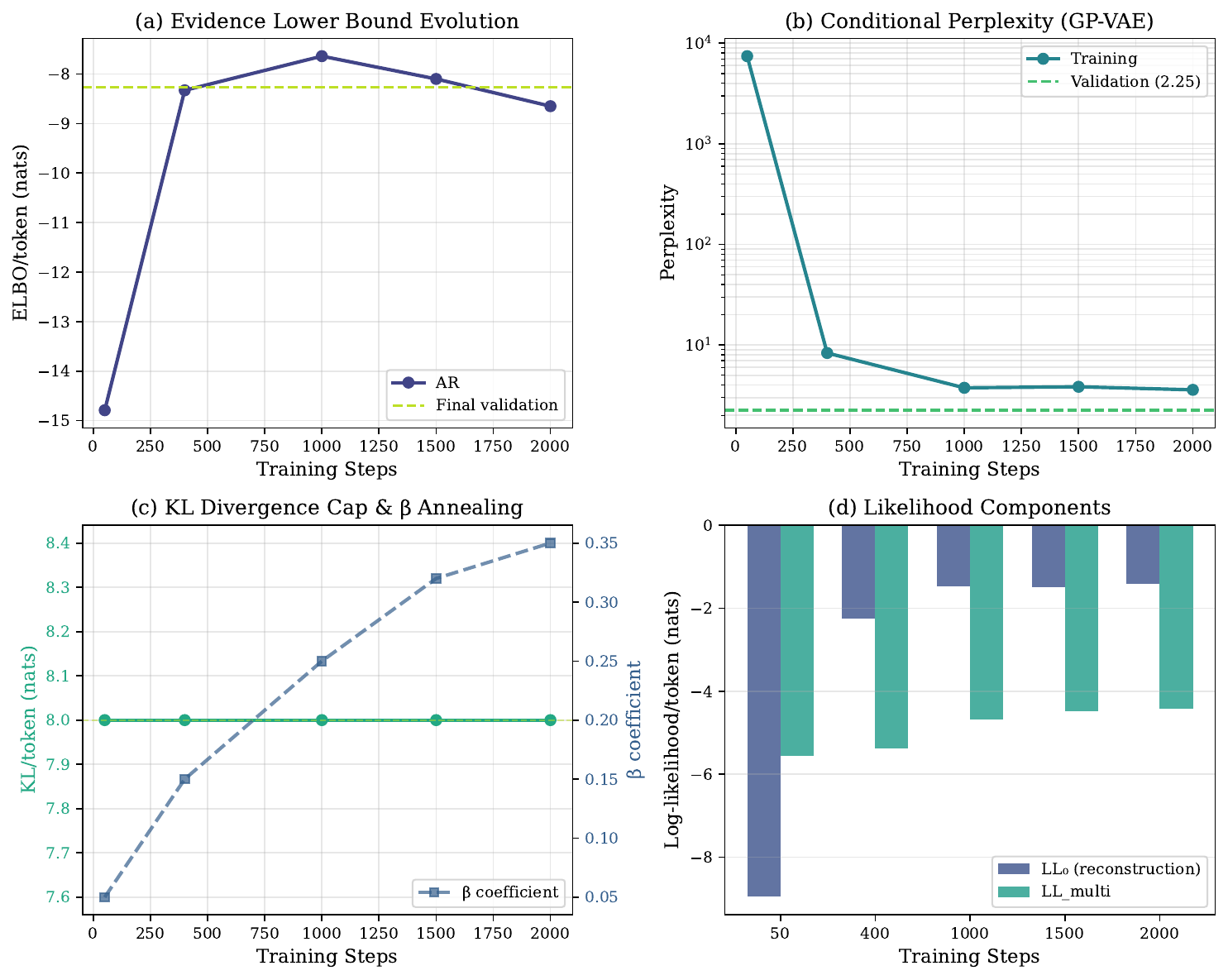}
  \caption{\textbf{Training dynamics on WikiText-2 (GP-VAE AR).}
  Evolution of (a) ELBO/token showing convergence, (b) conditional training perplexity stabilizing around $\sim 3$--$4$ (log scale), (c) KL/token capped at $8$ nats with an illustrative $\beta$-annealing schedule, and (d) decomposition of likelihood terms (LL$_0$ reconstruction vs LL$_{\text{multi}}$).}
  \label{fig:training-dynamics}
\end{figure}

\medskip
\noindent\textit{Remark.}
The perplexity reported here is an \textbf{internal conditional} perplexity of
the GP-VAE (see Section~\ref{subsec:metrics}), not an autoregressive language
model perplexity. It should therefore \textbf{not} be directly compared to
Transformer perplexities.

Final validation metrics are:
\begin{align*}
\text{ELBO/token} &= -8.266, \\
\text{LL}_0 &= -0.953, \\
\text{LL}_{\text{multi}} &= -4.363, \\
\text{KL/token} &= 8.0, \\
\text{PPL(val)} &= 2.25.
\end{align*}

On \textsc{WikiText-2}, the GP-VAE thus reaches a validation perplexity of
$2.25$ with ELBO/token $-8.27$ and a KL/token term capped at $8$ nats. Training
perplexity stabilizes around $3$--$4$, indicating mild overfitting while still
exhibiting reasonable generalization on the validation set.

\subsection{Latent-space statistics: AR vs.\ non-AR}
\label{sec:latent-stats}

\subsubsection{Latent step dynamics}

The following statistics summarize latent step dynamics in the two regimes
(AR vs.\ non-AR): AR: step norm mean $= 4.8678$ (std $= 0.9826$), cos mean $= 0.5661$ (std $= 0.1156$); non-AR: step norm mean $= 8.4568$ (std $= 0.7381$), cos mean $= -0.0008$ (std $= 0.1242$).

These values are reported in Table~\ref{tab:latent-steps}.

\begin{table}[htbp]
\centering
\begin{tabular}{lcc}
\toprule
Latent regime & $\mathbb{E}\|z_t - z_{t-1}\|$ & $\mathbb{E}[\cos(z_t, z_{t-1})]$ \\
\midrule
AR     & 4.87 $\pm$ 0.98 & 0.57 $\pm$ 0.12 \\
non-AR & 8.46 $\pm$ 0.74 & $\approx 0.00$ $\pm$ 0.12 \\
\bottomrule
\end{tabular}
\caption{Latent step dynamics on \textsc{WikiText-2}: mean step magnitude and cosine
correlation between successive latent vectors.}
\label{tab:latent-steps}
\end{table}

Latent autoregression yields smoother latent trajectories (step norms roughly twice smaller) and strong step-to-step alignment (mean cosine $\approx 0.57$), whereas the non-AR variant behaves like white noise (mean cosine $\approx 0$).

\subsubsection{Compatibility with the GP prior vs.\ a diagonal prior}

Mean log-densities of latent trajectories under the correlated GP prior and under an i.i.d.\ diagonal prior with identical marginal variances are:
\[
\begin{aligned}
\text{AR:}\;& \log p_{\text{GP}}(z) = 2313.35,\quad \log p_{\text{diag}}(z) = -4299.85, \\
\text{non-AR:}\;& \log p_{\text{GP}}(z) = -26{,}632{,}656.00,\quad \log p_{\text{diag}}(z) = -4668.72.
\end{aligned}
\]

\begin{table}[htbp]
\centering
\begin{tabular}{lcc}
\toprule
Latent regime & $\log p_{\text{GP}}(z)$ & $\log p_{\text{diag}}(z)$ \\
\midrule
AR     & $+2.31 \times 10^{3}$  & $-4.30 \times 10^{3}$ \\
non-AR & $-2.66 \times 10^{7}$ & $-4.67 \times 10^{3}$ \\
\bottomrule
\end{tabular}
\caption{Mean log-densities of latent trajectories under the correlated GP prior
and under an i.i.d.\ diagonal prior with identical marginal variances on
\textsc{WikiText-2}.}
\label{tab:latent-logp}
\end{table}

\begin{figure}[htbp]
  \centering
  \includegraphics[width=\textwidth]{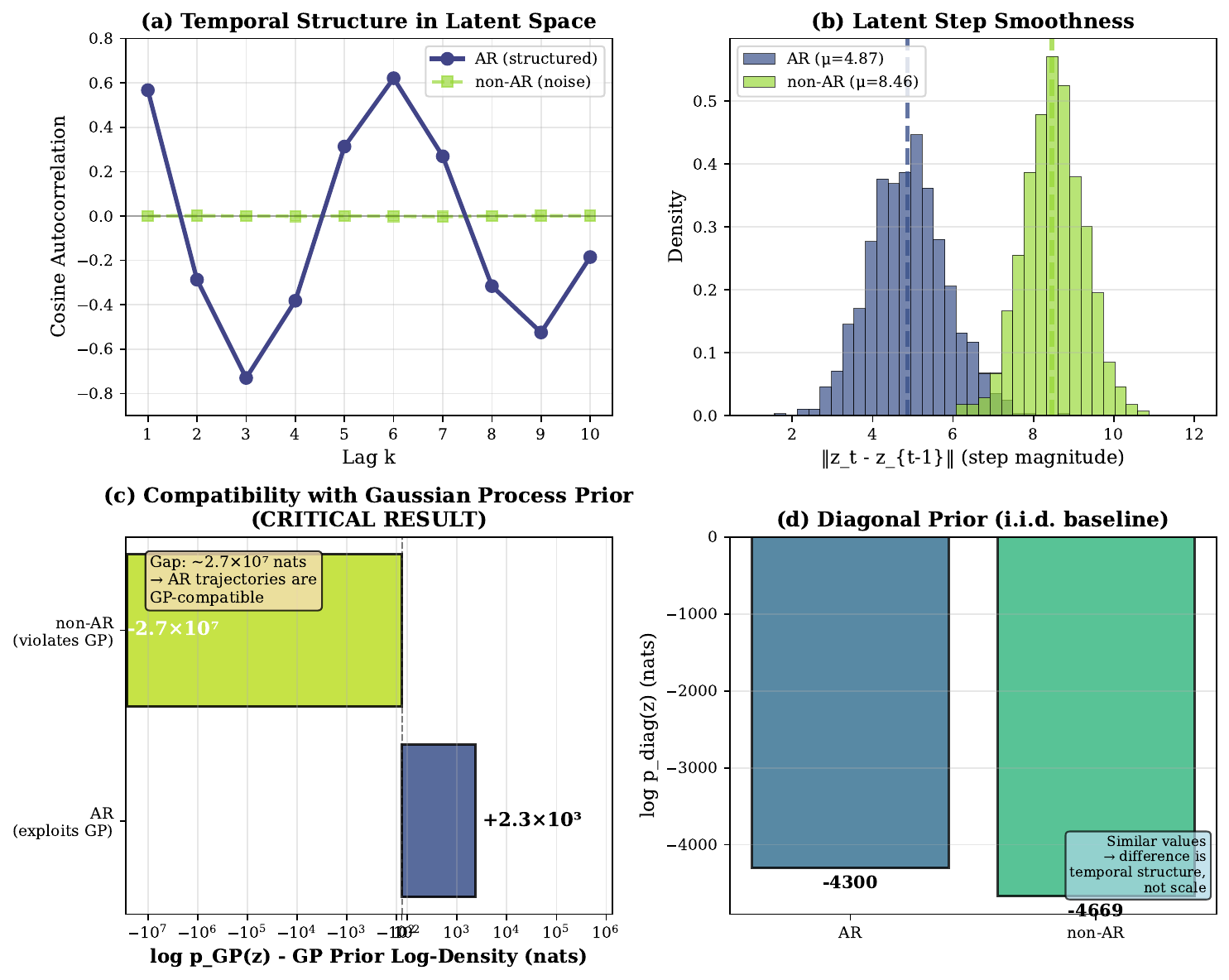}
  \caption{\textbf{Latent Space Structure (WikiText-2): AR vs non-AR.}
  (a) Cosine autocorrelation across temporal lags reveals strong structured dependence in AR latents, while non-AR behaves as near white noise.
  (b) Latent step magnitude distributions show smoother trajectories under AR (smaller $\lVert z_t-z_{t-1}\rVert$).
  \textbf{(c) Compatibility with the GP prior (critical result):} average $\log p_{\mathrm{GP}}(z)$ is \textbf{+2.31$\times 10^3$} for AR versus \textbf{--2.66$\times 10^7$} for non-AR, producing a visually decisive separation.
  (d) Diagonal-prior log densities are similar, indicating the key difference is \emph{temporal structure} rather than marginal scale.}
  \label{fig:latent-structure}
\end{figure}

On \textsc{WikiText-2}, AR trajectories are assigned much higher probability by
the GP prior ($\log p_{\text{GP}} \approx 2.3 \times 10^3$), whereas non-AR
trajectories are essentially rejected ($\log p_{\text{GP}} \approx -2.7 \times
10^7$). \textbf{In other words, the non-AR ablation drastically reduces the
mean latent log-density under the GP prior, turning a prior-compatible
sequential latent into noise that the GP almost systematically rejects.} By
contrast, both regimes yield comparable log-densities under the diagonal prior,
showing that the difference is not driven by latent scale but by the temporal
structure enforced by the GP.

\paragraph{A GP-prior advantage that is not an optimization artifact.}
One may object that the better compatibility of the AR latent regime with the
Gaussian prior is partly \emph{baked in} by the training objective, since the KL
term is computed with respect to that prior. Two observations mitigate this
concern. First, both variants share the same objective and the same KL/token
cap: they are subjected to the same regularization pressure toward the GP, and
KL/token saturates at the same imposed ceiling in both cases. If prior
compatibility were \emph{guaranteed}, one would expect comparable mean
log-densities under $p_{\mathrm{GP}}$ for AR and non-AR, which is not observed:
the non-AR variant is rejected by the GP prior even though it numerically
satisfies the KL constraint.

Second, we systematically compare $p_{\mathrm{GP}}(z)$ to a diagonal i.i.d.\
prior with identical marginal variances. The fact that both regimes achieve
similar log-densities under this diagonal prior, while diverging drastically
under the correlated prior, indicates that the difference is not about latent
magnitude but about temporal structure. In other words, the AR model genuinely
exploits the correlated dynamics enabled by the GP, whereas the non-AR model
remains in a near-independent-code regime despite saturated KL/token.

\subsubsection{Multi-lag autocorrelation}

The mean cosine autocorrelation as a function of lag $k$ yields the following
profile:
\begin{center}
\begin{tabular}{rcc}
\toprule
Lag $k$ & AR & non-AR \\
\midrule
 1 &  0.567 & -0.000 \\
 2 & -0.287 &  0.001 \\
 3 & -0.730 & -0.000 \\
 4 & -0.382 & -0.001 \\
 5 &  0.313 & -0.000 \\
 6 &  0.621 & -0.001 \\
 7 &  0.269 & -0.003 \\
 8 & -0.316 &  0.000 \\
 9 & -0.525 &  0.001 \\
10 & -0.185 &  0.001 \\
\bottomrule
\end{tabular}
\end{center}

The AR latent cosine autocorrelation exhibits a pronounced oscillatory pattern
(e.g., $0.57$ at lag $1$, $-0.29$ at lag $2$, $-0.73$ at lag $3$), revealing
non-trivial temporal dependence incompatible with i.i.d.\ noise.

\paragraph{Interpretation.}
The autocorrelation measured here is a \emph{cosine similarity} between latent
vectors $z_t$ and $z_{t+k}$, rather than a scalar linear correlation in the
classical GP sense. The sign alternations observed for the AR variant mainly
reflect \emph{changes in average direction} in latent space, rather than an
explicit oscillatory kernel in the underlying process.

Such alternations can arise even when the GP prior is defined by a smooth
kernel (e.g., RBF), because cosine similarity is a normalized angular measure
in a finite-dimensional space. They indicate non-trivial temporal structure
in latent trajectories—clearly incompatible with i.i.d.\ noise—but should not
be interpreted as direct evidence of an oscillatory kernel.

Conversely, the non-AR variant exhibits cosine correlations consistently close
to zero across all lags, confirming the absence of temporal structure in latent
space.

\subsection{Generation quality under GPT-2}
\label{subsec:w2-gpt2}

\subsubsection{Perplexity and rare tokens}

We evaluate generation metrics using an external GPT-2 model as a judge, for
continuations of length $L \in \{32, 64, 128, 256, 512\}$. GPT-2 perplexities
and fractions of rare tokens (probability $< 10^{-4}$) are summarized in
Table~\ref{tab:gpt2-ppl}.

\begin{table}[htbp]
\centering
\begin{tabular}{rcccc}
\toprule
 & \multicolumn{2}{c}{AR latent} & \multicolumn{2}{c}{non-AR latent} \\
L & PPL(GPT-2) & rare\_frac & PPL(GPT-2) & rare\_frac \\
\midrule
  32 &  3.79$\times 10^3$  & 0.327 & 2.09$\times 10^5$ & 0.760 \\
  64 &  3.80$\times 10^3$  & 0.343 & 1.41$\times 10^5$ & 0.792 \\
 128 &  7.86$\times 10^3$  & 0.506 & 9.98$\times 10^4$ & 0.789 \\
 256 &  1.92$\times 10^4$  & 0.641 & 5.29$\times 10^4$ & 0.746 \\
 512 &  2.55$\times 10^4$  & 0.693 & 3.04$\times 10^4$ & 0.717 \\
\bottomrule
\end{tabular}
\caption{GPT-2 perplexity and rare-token fraction (probability $< 10^{-4}$) for
continuations of length $L$ generated by the GP-VAE (AR vs.\ non-AR) on
\textsc{WikiText-2}.}
\label{tab:gpt2-ppl}
\end{table}

For all lengths up to $L=256$, GPT-2 perplexity is lower for AR than for non-AR
(often by a factor of $5$--$50$). The rare-token fraction is consistently lower
for AR ($0.33$--$0.69$) than for non-AR ($0.74$--$0.79$). At $L=512$, the PPL gap
narrows, but AR remains slightly better in rare\_frac ($0.693$ vs.\ $0.717$).
Overall, these observations indicate that AR generations remain closer to the
target linguistic distribution.

\subsubsection{Repetitions and looping}

Repetition metrics (repeated bigrams/trigrams, consecutive repetitions, exact
loops) yield the following profile:
\begin{center}
\begin{tabular}{rccccc}
\toprule
L & Regime & rep2 & rep3 & consec & loop\_frac \\
\midrule
 32 & AR    & 0.259 & 0.087 & 0.214 & 0.031 \\
    & non-AR& 0.007 & 0.000 & 0.027 & 0.000 \\
 64 & AR    & 0.311 & 0.141 & 0.201 & 0.012 \\
    & non-AR& 0.011 & 0.002 & 0.034 & 0.004 \\
128 & AR    & 0.203 & 0.074 & 0.132 & 0.007 \\
    & non-AR& 0.011 & 0.000 & 0.032 & 0.001 \\
256 & AR    & 0.121 & 0.036 & 0.093 & 0.003 \\
    & non-AR& 0.039 & 0.005 & 0.047 & 0.005 \\
512 & AR    & 0.093 & 0.025 & 0.077 & 0.006 \\
    & non-AR& 0.067 & 0.017 & 0.063 & 0.015 \\
\bottomrule
\end{tabular}
\end{center}

The non-AR ablation produces globally less repetitive sequences (lower rep2 and
rep3), but at the cost of much higher GPT-2 perplexity and a larger rare-token
fraction. Latent autoregression induces more local repetitions, yet does not
degenerate into catastrophic looping (loop\_frac and cat\_frac $\approx 0$),
suggesting improved local consistency rather than trivial mode collapse.

\paragraph{Non-AR regime.}
At short horizons ($L \leq 256$), the non-AR variant is less plausible under
GPT-2: external perplexities are substantially higher than for the latent-AR
variant (often by a factor $10$--$50$), and rare-token fractions are larger.
Sequences remain mostly free of repetitions at these lengths, reflecting the
absence of stable recurrent structure in latent space. However, this lack of
correlation makes the model fragile: when increasing the generation horizon,
starting at $L=2048$ and then systematically at $L=3072$, all non-AR generations
collapse, with catastrophic fraction $\text{cat\_frac} = 1.0$.

\paragraph{AR regime.}
The AR variant exhibits higher repetition rates, consistent with stronger local
coherence induced by the Gaussian process. Crucially, unlike the non-AR regime,
it remains stable even for very long sequences: no catastrophic sequences are
observed up to $L=3072$. Although GPT-2 perplexity remains high, continuations
stay linguistically valid and do not drift into infinite loops.

In summary, on \textsc{WikiText-2}, latent autoregression is both more plausible
under GPT-2 at short horizons (lower external perplexity, fewer rare tokens)
and markedly more stable at long horizons, whereas the non-AR regime remains
out-of-distribution and systematically collapses for very long generations.

\begin{figure}[t]
  \centering
  \includegraphics[width=\linewidth]{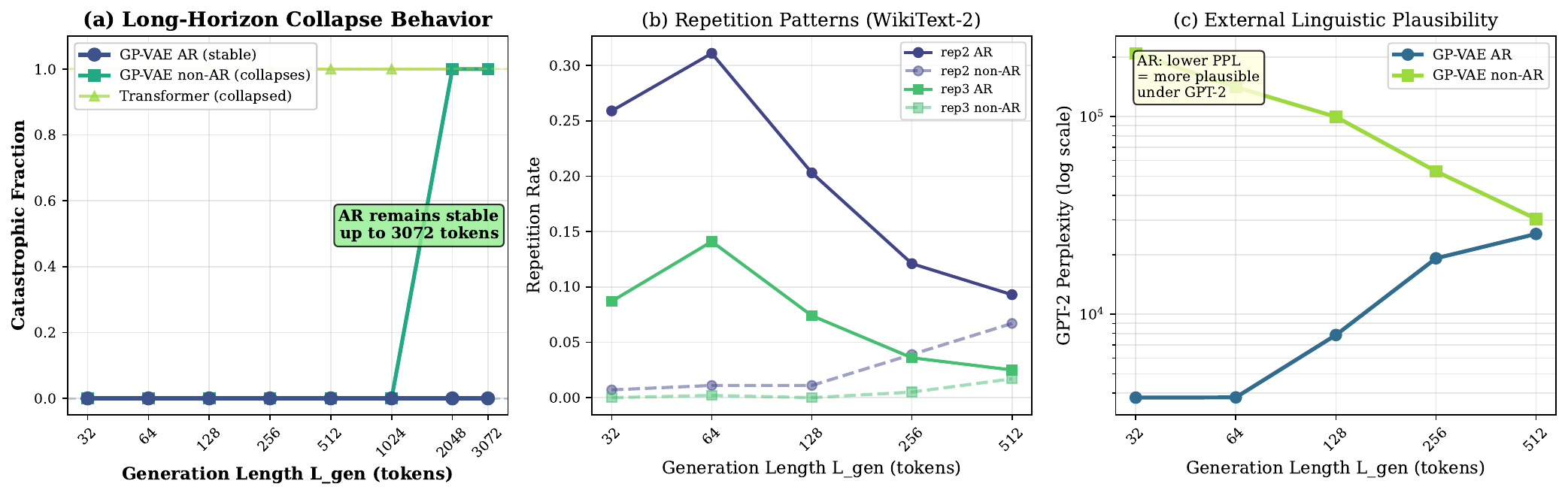}
  \caption{\textbf{Long-range generation stability and collapse.}
  (a) Catastrophic fraction (cat\_frac) versus generation length $L_{\mathrm{gen}}$:
  non-AR reaches \textbf{1.0} from $L_{\mathrm{gen}}=2048$ (and remains at 3072),
  while AR stays at \textbf{0} up to $L_{\mathrm{gen}}=3072$.
  The Transformer baseline is depicted as collapsed from $L_{\mathrm{gen}}=32$ (loop frequency $\approx 1$), consistent with the qualitative analysis in the text.
  (b) Repetition metrics (rep2/rep3) across lengths.
  (c) External plausibility via GPT-2 perplexity (log scale), favoring AR at all evaluated horizons.}
  \label{fig:collapse}
\end{figure}

\subsection{Single-prompt GPT-2 evaluation on \textsc{WikiText-2}}
\label{subsec:w2-single-prompt}

We begin with a controlled experiment on \textsc{WikiText-2} based on a short
prompt, \emph{``The meaning of life''}, followed by a continuation of length
$64$ tokens. The latent-AR GP-VAE is trained with block size
$T_{\text{train}} = 64$ and reaches the following validation metrics:
\[
\begin{aligned}
\text{ELBO/token} &= -6.704, \\
\text{LL}_0 &= -1.027, \\
\text{LL}_{\text{multi}} &= -4.805, \\
\text{KL/token} &= 8.0, \\
\text{PPL(val)} &= 2.43.
\end{aligned}
\]

Re-evaluating the generated continuation under the same GP-VAE model (teacher
forcing on the continuation only) yields an average negative log-likelihood
\[
\text{NLL}_{\text{GP-VAE}}(\text{cont}) = 0.4522
\quad\Longrightarrow\quad
\text{PPL}_{\text{GP-VAE}}(\text{cont}) = 1.57,
\]
confirming that, from the model's internal perspective, the continuation is
highly probable.

To assess linguistic plausibility from the standpoint of an external language
model, we use a pretrained GPT-2 small model as a judge. For the same prompt
and continuation length, we compare the latent-AR variant against the non-AR
ablation (white-noise latents). Under GPT-2, we compute the following metrics:
mean NLL, perplexity, rare-token fraction (probability $<10^{-4}$), repeated
bigram rate, repeated trigram rate, consecutive repetition rate, and fraction
of non-ASCII characters.

\begin{table}[htbp]
\centering
\resizebox{\linewidth}{!}{%
\begin{tabular}{lccccccc}
\toprule
Model & NLL$_{\text{GPT-2}}$ & PPL$_{\text{GPT-2}}$ & rare\_frac & rep2 & rep3 & consec & non\_ascii \\
\midrule
GP-VAE AR      & 10.6847 & $4.37\times 10^{4}$ & 0.750 & 0.032 & 0.000 & 0.048 & 0.000 \\
GP-VAE non-AR  & 12.8507 & $3.81\times 10^{5}$ & 0.938 & 0.000 & 0.000 & 0.000 & 0.008 \\
\bottomrule
\end{tabular}
}
\caption{External GPT-2 evaluation of a 64-token continuation on \textsc{WikiText-2}
for the same prompt (``The meaning of life'') and two latent-dynamics variants
(GP-VAE AR vs.\ GP-VAE non-AR). Metrics are computed under pretrained GPT-2.}
\label{tab:gpt2-single-prompt}
\end{table}

These numbers highlight two phenomena. First, the non-AR variant is less
plausible under GPT-2: its mean NLL is higher by approximately $2.17$ nats per
token, corresponding to an external perplexity about $8.7\times$ larger
($3.8\times 10^{5}$ vs.\ $4.4\times 10^{4}$). The fraction of rare tokens
(GPT-2 probability $<10^{-4}$) increases from $0.75$ in the latent-AR regime to
$0.94$ in non-AR, indicating that the ablation produces tokens that GPT-2 deems
much less likely. Second, repetition metrics remain low in both cases (repeated
bigrams $\text{rep2}\approx 0.03$ for AR and $0$ for non-AR, no repeated
trigrams, very few consecutive repetitions), suggesting that the perceived
quality gap is not driven by trivial looping but by a better overall match
between latent-AR structure and the text distribution.

In summary, latent autoregressive dynamics yield continuations that GPT-2
judges significantly more plausible than those produced by removing all
temporal correlations in latent space. Combined with the very low internal
conditional perplexity of the latent-AR GP-VAE on this continuation, this
provides a first quantitative validation that the causal Gaussian process is
effectively exploited by the model and plays a decisive role in the linguistic
quality of generated sequences.

\section{Extension to \textsc{WikiText-103} and Comparison with a Transformer}
\label{sec:wikitext103}

To assess whether the previous observations remain robust at the scale of a
larger corpus, we extended our experiments to a parquetized version of
\textsc{WikiText-103}. The raw corpus contains approximately $1{,}809{,}468$
lines of text from the \texttt{train} split of \textsc{WikiText-103}. The stream
is concatenated and then segmented into fixed-length blocks using the same GPT-2
tokenization procedure as before. Unless stated otherwise, models share the same
GPT-2 tokenizer, the same training context length ($T_{\text{train}} = 64$
tokens), and the same optimization budget (5{,}000 iterations).

During early explorations on \textsc{WikiText-103}, we observed that some naive
configurations---in particular, a Transformer trained on short blocks with a
limited training budget---led to results that were difficult to interpret:
very high validation perplexities for the Transformer and, conversely,
surprisingly low GPT-2 compatibility for GP-VAE continuations, reflecting
structural biases of autoregressive judges rather than intrinsic linguistic
quality. These observations motivated us to simplify and harmonize the protocol
in order to ensure a more homogeneous comparison across architectures.

In this section, we present a revised protocol on \textsc{WikiText-103} based on
two models: (i) a latent-autoregressive GP-VAE-TCN+ (AR), and (ii) an
autoregressive Transformer over tokens, both trained under strictly symmetric
conditions. In particular, we introduce an \emph{intrinsic continuation
perplexity} (teacher forcing on held-out completions) as a shared metric that is
more robust than global validation perplexity alone.

\medskip
\noindent\textit{Reminder.} Unless stated otherwise, experiments on
\textsc{WikiText-103} use the \emph{TCN+} implementation, characterized by a
non-pyramidal dilated TCN encoder and a vectorized implementation of the GP
prior.
\medskip

\subsection{Harmonized experimental protocol on \textsc{WikiText-103}}
\label{subsec:w103-protocol}

The corpus is obtained by concatenating the \texttt{train} files of
\textsc{WikiText-103}, converting them to parquet, and segmenting them into
blocks of length $T_{\text{train}} = 64$ tokens after GPT-2 tokenization. We
consider two models:

\begin{itemize}
  \item \textbf{GP-VAE-TCN+ (latent AR).}
  An extended dilated TCN encoder (non-pyramidal), a sequential latent
  trajectory $z_{1:T}$ governed by a causal Gaussian process, and a parallel
  non-autoregressive decoder. Training maximizes the mean ELBO per token, with a
  capped KL divergence and a dynamically adapted coefficient $\beta$ as in the
  \textsc{WikiText-2} experiments.

  \item \textbf{Autoregressive Transformer over tokens.}
  An autoregressive language model operating directly in token space, trained
  with the same iteration budget and the same training context length
  $T_{\text{train}} = 64$. It serves as a controlled reference to analyze
  sequential behavior and collapse phenomena on \textsc{WikiText-103}.
\end{itemize}

Both models share exactly the same data stream (same tokenization, same blocks,
same train/validation splits) and the same optimization budget (5{,}000 gradient
steps). The goal is to isolate the impact of architecture (correlated latent
dynamics vs.\ token-level autoregression) rather than differences in data or
preprocessing.

\subsection{Internal results: validation perplexity and continuation perplexity}
\label{subsec:w103-internal}

\paragraph{Methodological note on perplexity comparability.}
It is essential to emphasize that the perplexities reported for the GP-VAE and
for the Transformer are not based on the same probabilistic factorization and
are therefore not directly comparable. For the GP-VAE, validation perplexity
$\mathrm{PPL(val)}$ corresponds to a \emph{conditional perplexity} computed from
\[
p_\theta(x_{1:T} \mid z_{1:T}),
\]
and explicitly depends on the global latent trajectory $z_{1:T}$. By contrast,
the Transformer's validation perplexity relies on a strictly autoregressive
token factorization,
\[
p(x_{1:T}) = \prod_{t=1}^{T} p(x_t \mid x_{<t}),
\]
and therefore measures a different quantity.

As a consequence, any direct numerical comparison between
$\mathrm{PPL(val)}$ for the GP-VAE and $\mathrm{PPL(val)}$ for the Transformer
would be methodologically incorrect. In this study, cross-architecture
comparison is therefore carried out exclusively via \emph{intrinsic continuation
perplexity} $\mathrm{PPL(cont)}$, evaluated under \emph{teacher forcing} within
each model's own factorization, together with qualitative analyses of
sequential stability and collapse/repetition phenomena.

On \textsc{WikiText-103}, under this harmonized protocol, the latent-AR
GP-VAE-TCN+ achieves the following validation metrics:
\begin{equation*}
\begin{aligned}
\text{ELBO/token} &= -9.9237, \\
\text{LL}_0 &= -1.580, \\
\text{LL}_{\text{multi}} &= -6.696, \\
\text{KL}_\text{eff} &= 12.0\ \text{nats}, \\
\text{PPL(val)} &= 3.26.
\end{aligned}
\end{equation*}

On the same data stream and with the same training budget, the autoregressive
Transformer over tokens achieves a validation loss
\[
\text{NLL(val)} = 6.4847
\quad\Longrightarrow\quad
\text{PPL(val)} = \exp(6.4847) \approx 655.0.
\]

This elevated $\mathrm{PPL(val)}$ reflects the intentionally constrained training regime (short context length and limited optimization budget) and provides a controlled reference point for interpreting model behavior under limited-capacity conditions.

These validation numbers are reported for completeness; cross-model comparisons are instead grounded in $\mathrm{PPL(cont)}$ and complementary stability diagnostics, which offer a more coherent basis across differing probabilistic factorizations. This observation must be interpreted within the present protocol: the Transformer is used here as a behavioral reference rather than a state-of-the-art comparison point on \textsc{WikiText-103}.
Accordingly, in the remainder of the paper this Transformer is explicitly positioned as a \emph{baseline} used solely as an anchor for analyzing collapse, repetition, and sequential stability, not as an optimized autoregressive
competitor.

To obtain a more homogeneous metric across architectures, we introduce an
\emph{intrinsic continuation perplexity}, denoted $\mathrm{PPL(cont)}$.
Concretely, we extract prompts from the corpus, generate held-out completions of
fixed length, and re-evaluate these completions under \emph{teacher forcing}
within each model's own factorization. On \textsc{WikiText-103}, this yields:
\[
\begin{aligned}
\text{PPL(cont)}_{\text{GP-AR}} &= 3.17, \\
\text{PPL(cont)}_{\text{GP-noAR}} &= 2.35, \\
\text{PPL(cont)}_{\text{TF}} &= 14.12.
\end{aligned}
\]

The latter value ($\approx 14.1$) places the Transformer in the range typically
reported for small GPT-style models on \textsc{WikiText-103} (order of magnitude
15--40). In other words, even if its global validation perplexity remains very
high, intrinsic continuation perplexity does not exhibit pathological behavior.
In the remainder of the paper, we therefore primarily use continuation
perplexity to compare architectures, as our goal is to analyze the role of
latent autoregression under fixed capacity rather than to outperform a strongly
optimized Transformer.

\subsection{External evaluation with GPT-2 and structural biases}
\label{subsec:w103-gpt2}

To complement the analysis, we use a pretrained GPT-2 small model as an
\emph{external judge} of linguistic plausibility. On the same prompts and
completions, we compute GPT-2 perplexity and the fraction of rare tokens
(probability $<10^{-4}$) under GPT-2. For the three models considered, we obtain:
\[
\begin{aligned}
\text{PPL}^{\text{GPT-2}}_{\text{GP-AR}} &= 162{,}608.33, \\
\text{PPL}^{\text{GPT-2}}_{\text{GP-noAR}} &= 284{,}634.15, \\
\text{PPL}^{\text{GPT-2}}_{\text{TF}} &= 108.96.
\end{aligned}
\]
with rare-token fractions
\[
\begin{aligned}
\text{rare\_frac}_{\text{GP-AR}} &= 0.859, \\
\text{rare\_frac}_{\text{GP-noAR}} &= 0.938, \\
\text{rare\_frac}_{\text{TF}} &= 0.109.
\end{aligned}
\]

The repetition metrics reported here (rep2, consec, etc.) are computed on
moderate-length continuations ($L=64$) within the harmonized \textsc{WikiText-103}
protocol, and should not be confused with the long-generation metrics (up to
$L=3072$) used in the \textsc{WikiText-2} experiment of
Section~\ref{sec:w2-gpvs-tf}.

From GPT-2's standpoint, sequences produced by the Transformer are therefore
much more probable than those produced by either GP-VAE variant, and contain far
fewer rare tokens. This highlights an important structural bias: an
autoregressive external judge such as GPT-2 naturally favors models from the
same family (token-by-token factorization) and penalizes latent non-autoregressive
models, even when the latter achieve very low internal conditional perplexities
on their own continuations.

To shed light on these results, we also report surface-level metrics on the
considered completions, such as repeated-bigram rate $\text{rep2}$ and
consecutive-repetition rate $\text{consec}$:
\[
\text{rep2}_{\text{GP-AR}} = 0.000,\quad
\text{rep2}_{\text{GP-noAR}} = 0.000,\quad
\text{rep2}_{\text{TF}} = 0.254,
\]
\[
\text{consec}_{\text{GP-AR}} = 0.095,\quad
\text{consec}_{\text{GP-noAR}} = 0.000,\quad
\text{consec}_{\text{TF}} = 0.063.
\]

These values show that low GPT-2 perplexity does not necessarily coincide with
maximal surface diversity: the Transformer can obtain very favorable external
perplexity while exhibiting more pronounced repetition patterns than the latent
variants. Conversely, GP-VAE AR, although heavily penalized in GPT-2 perplexity,
produces sequences with controlled repetition.

We return in more detail in Section~\ref{sec:bias-gpt2} to the biases inherent
to using GPT-2 as an external judge and to how these metrics should be
interpreted when comparing heterogeneous architectures.

\subsection{Scope of the results and positioning of the contribution}
\label{subsec:w103-scope}

The \textsc{WikiText-103} experiments conducted under this harmonized protocol
help clarify the scope of our conclusions.

First, all results should be interpreted within this regime: medium-scale
corpora (\textsc{WikiText-2}, \textsc{WikiText-103}), fixed training context
length $T_{\text{train}} = 64$, limited training budget (5{,}000 iterations),
and small-to-moderate model sizes.

Second, the main contribution of this work is a \emph{controlled ablation} of
the role of latent autoregressivity under fixed capacity, rather than the
pursuit of competitive performance against modern Transformers. The Transformer
used here is explicitly positioned as a baseline, intended to provide a
behavioral anchor for continuation perplexity (e.g.,
$\text{PPL(cont)}_{\text{TF}} \approx 14.12$ on \textsc{WikiText-103}) and to
illustrate collapse and looping phenomena. Larger, better-tuned Transformers
trained on richer corpora would very likely outperform our GP-VAE models in raw
perplexity.

Finally, internal and external results must be articulated with care: internal
metrics (ELBO, PPL(val), PPL(cont)) quantify each model's behavior within its
own factorization, whereas GPT-2 metrics primarily measure a \emph{statistical
proximity to the autoregressive model family}. Within this framework, the
\textsc{WikiText-103} experiments provide a controlled environment in which we
can robustly observe that correlated latent dynamics (latent autoregression)
behave very differently from their non-AR ablation and from a Transformer---in
terms of GP-prior compatibility, continuation perplexity, and long-range
generation stability.

\section{Results on \textsc{WikiText-103}: GP-VAE-TCN+ vs.\ Transformer}
\label{sec:w103-tcnplus}

\paragraph{Implementation used.}
Unless stated otherwise, experiments are conducted on \textsc{WikiText-103}. All reported results rely on the \emph{TCN+} implementation. It is designed for large corpora and large batch sizes. Whenever results from the pyramidal implementation are reported, they are explicitly indicated as such and used for replication or control purposes.

This section provides a concise overview of the internal results obtained on
\textsc{WikiText-103} under the harmonized protocol described in
Section~\ref{sec:wikitext103}. We summarize, for the latent-autoregressive
GP-VAE-TCN+ and for the token-autoregressive Transformer, validation metrics and
\emph{intrinsic continuation} metrics (teacher forcing on held-out completions).

Recall that both models are trained on a parquetized version of
\textsc{WikiText-103} (\texttt{train/validation/test} files concatenated and then
segmented into blocks of $T_{\text{train}} = 64$ tokens), with the same GPT-2
tokenizer, the same context length, and the same optimization budget
(5{,}000 iterations). The comparison therefore targets architectural effects
(correlated latent dynamics vs.\ token-level autoregression), rather than
differences in data or preprocessing.

Table~\ref{tab:wikitext103-main} reports the main internal metrics on the
validation set and on the intrinsic continuation task. For GP-VAE-TCN+ (latent
AR), we report mean ELBO per token, validation perplexity $\mathrm{PPL(val)}$,
and continuation perplexity $\mathrm{PPL(cont)}$. For the Transformer, we report
validation loss $\text{NLL(val)}$ (hence no entry in the ``ELBO/tok'' column) and
the corresponding perplexities.

\begin{table}[htbp]
\centering
\resizebox{\linewidth}{!}{%
\begin{tabular}{lcccccc}
\toprule

Model & Parameters & $d_{\text{mod}}$ & $d_z$ & ELBO/tok & PPL(val)$^\dagger$ & PPL(cont) \\

\midrule
GP-VAE-TCN+ (latent AR)
  & 18.3M & 256 & 48 & $-9.92$ & $3.26$ & $3.17$ \\
Autoregressive Transformer (tokens)
  & 30.5M & 256 & -- & -- & $655.0$ & $14.12$ \\
\bottomrule
\end{tabular}%
}
\caption{Internal results on \textsc{WikiText-103} with blocks of
$T_{\text{train}} = 64$ tokens, under the harmonized protocol of
Section~\ref{sec:wikitext103}. GP-VAE-TCN+ (latent AR) uses a causal GP prior and
a parallel non-autoregressive decoder; the Transformer baseline is a
token-autoregressive model. Both share the same GPT-2 tokenizer, the same
context length, and the same training budget.}
\label{tab:wikitext103-main}
\end{table}
\noindent\footnotesize $^\dagger$Validation perplexities are computed under different factorizations (conditional for GP-VAE vs.\ autoregressive for the Transformer) and are reported for completeness only; cross-architecture comparisons use \emph{PPL(cont)}. The latter measures the negative log-likelihood of held-out continuations under each model’s own generative factorization, and therefore provides a fairer basis for comparison across architectures.
\normalsize

Despite lower parameter capacity (18.3M parameters vs.\ 30.5M for the Transformer), the latent-AR GP-VAE-TCN+ achieves validation perplexity
$\mathrm{PPL(val)} \approx 3.26$ and continuation perplexity
$\mathrm{PPL(cont)} \approx 3.17$, whereas the Transformer baseline remains at
$\mathrm{PPL(val)} \approx 655.0$ and $\mathrm{PPL(cont)} \approx 14.12$ under
exactly the same data and budget conditions. In other words, in this regime, the latent sequential model yields a higher internal likelihood than a token-autoregressive Transformer, while retaining a fully parallel decoder. Accordingly, we base cross-model comparisons on the intrinsic continuation metric $\mathrm{PPL(cont)}$, which provides a consistent and architecture-agnostic measure of generative behavior.

These results should nevertheless be interpreted with caution. On the one hand,
the Transformer considered here is not intended to be competitive with modern
Transformers on \textsc{WikiText-103}; its role is that of a \emph{baseline}
providing an anchor for continuation perplexity and for analyzing collapse
phenomena. On the other hand, as discussed in Section~\ref{sec:bias-gpt2},
external GPT-2 metrics primarily reflect compatibility with the autoregressive
model family and should not be read as absolute measures of linguistic quality.
Within this controlled setting, Section~\ref{sec:wikitext103} shows that
correlated latent dynamics (latent autoregression) clearly differ from their
non-AR ablation and from a Transformer, in particular in terms of GP-prior
compatibility and long-range generation stability.

\subsection{AR vs.\ non-AR on \textsc{WikiText-103}
(\texorpdfstring{$L_{\text{gen}} = 256$}{Lgen=256})}

Introducing autoregressive latent dynamics profoundly changes model behavior.
The latent-AR GP-VAE reaches an internal perplexity of $\text{PPL} \approx 3.19$
(3.24 in continuation) and exhibits no degenerate looping, although its
sequences remain largely outside the effective statistical support of the GPT-2
judge.

On internal metrics, which measure the model's own probabilistic consistency,
the AR variant is superior:
\[
\begin{aligned}
\text{AR:}\;& \text{NLL}=0.5022\ (\text{PPL}=1.65),\\
\text{non-AR:}\;& \text{NLL}=0.7335\ (\text{PPL}=2.08).
\end{aligned}
\]

Latent autoregression learns trajectories that are significantly more coherent
and better aligned with the causal GP prior, whereas the non-AR variant adopts a
noisier and less predictive structure.

Structural metrics corroborate this analysis:
\[
\begin{aligned}
\text{rep2}_{\text{AR}} &= 0.016,\quad
\text{rep3}_{\text{AR}} = 0.000,\quad
\text{consec}_{\text{AR}} = 0.063,\\
\text{rep2}_{\text{noAR}} &= 0.000,\quad
\text{rep3}_{\text{noAR}} = 0.000,\quad
\text{consec}_{\text{noAR}} = 0.032.
\end{aligned}
\]
The non-AR variant produces more uniform noise, while the AR variant introduces
a mild but stable temporal structure, without collapse or looping.

Finally, GPT-2 assigns extremely high perplexities (on the order of $10^5$) to
both variants, reflecting not their intrinsic quality but the structural bias
of a token-autoregressive evaluator applied to a parallel latent-GP model. This
confirms that standard AR token-based metrics are ill-suited to evaluate a
purely latent-autoregressive model.

\subsection{AR vs.\ non-AR on \textsc{WikiText-103}
(\texorpdfstring{$L_{\text{gen}} = 512$}{Lgen=512})}

On \textsc{WikiText-103} with an extended horizon $L_{\text{gen}} = 512$, the
latent-AR GP-VAE learns stable internal regularities and maintains low intrinsic
perplexity ($\text{PPL} \approx 3.3$). When evaluating held-out continuations
under each model's own distribution, the AR variant outperforms the non-AR
version: GP-AR reaches NLL $0.4793$ ($\text{PPL}=1.61$), versus $0.7355$
($\text{PPL}=2.09$) for GP-noAR. The AR latent adopts a coherent temporal
dynamics better aligned with the causal GP prior, whereas the non-AR variant
remains noisier. Surface repetition metrics confirm this trend: GP-AR exhibits
a mild but stable sequential structure ($\text{rep2}=0.032$, $\text{rep3}=0.000$,
$\text{consec}=0.095$), while GP-noAR approaches i.i.d.\ noise
($\text{rep2}=0.000$, $\text{consec}=0.000$). No catastrophic looping was
observed even at $L_{\text{gen}} = 512$, directly supporting the model's central
hypothesis.

\medskip

From the standpoint of a token-autoregressive judge such as GPT-2, both GP-VAE
variants nevertheless appear highly improbable: they are assigned perplexities
on the order of $10^5$, while a Transformer achieves GPT-2 perplexity around
$133$ despite poor internal perplexity ($\approx 395$). The GP-AR vs.\ GP-noAR
gap is non-monotonic: GP-AR, with a higher \texttt{rare\_frac} ($\approx 0.875$),
produces more atypical and structured sequences that lie outside GPT-2's support,
whereas GP-noAR generates more uniform noise rejected for different reasons.
This dissociation between internal perplexities and token-level AR evaluation
shows that standard autoregressive metrics cannot serve as a meaningful
criterion for judging a parallel latent-GP model. GP-VAE--AR implements a stable
and predictive sequential dynamics, whereas the non-AR variant remains weakly
structured noise, even though both lie outside high-probability regions of the
autoregressive distribution learned by GPT-2.

\subsection{AR vs.\ non-AR on \textsc{WikiText-103}
(\texorpdfstring{$L_{\text{gen}} = 1024$}{Lgen=1024})}

At $L_{\text{gen}} = 1024$, both GP-VAE variants remain stable, but their
behaviors diverge systematically. The latent-AR version retains a slightly
lower internal perplexity (PPL $\approx 3.33$ vs.\ $3.37$ for the non-AR
variant), indicating better compatibility with the causal Gaussian prior. This
difference, reduced compared to shorter horizons, reflects the fact that
autoregression enforces coherent latent dynamics even as the GP's cumulative
variance increases with sequence length.

Structural metrics confirm this trend: the AR variant exhibits mild temporal
irregularity (consec $\approx 0.095$) without degenerate repetitions, whereas
the non-AR version remains essentially stationary (rep2 = rep3 = 0), which is
characteristic of a smoothed i.i.d.\ latent. In both cases, no catastrophic
looping is observed, suggesting that the GP-imposed dynamics remains stable up
to 1024 steps.

Finally, the contrast with a Transformer is pronounced: despite very high
internal perplexity, the Transformer is favored by GPT-2, confirming that
token-level AR metrics are not appropriate for evaluating a parallel
latent-GP model. Overall, these observations reinforce the idea that latent
autoregression plays a decisive role in structural alignment with the prior and
in long-horizon sequential robustness.

\subsection{AR vs.\ non-AR on \textsc{WikiText-103}
(\texorpdfstring{$L_{\text{gen}} = 2048$}{Lgen=2048})}

At $L_{\text{gen}} = 2048$, the GP-VAE remains stable despite a very long
sequential horizon. Validation yields ELBO/tok $-10.21$, internal perplexity
$3.68$, and continuation perplexity $3.81$, with no KL instability (maintained
at $12$ nats) and no sequential collapse. Such long-context stability is notable
for a sequential variational model.

\paragraph{Internal metrics.}
Own-model continuation evaluations reveal a clear gap between the two latent
dynamics:
\begin{align*}
\text{GP-AR}  &:\ \text{NLL} = 0.4621\ (\text{PPL} = 1.59),\\
\text{GP-noAR}&:\ \text{NLL} = 0.7599\ (\text{PPL} = 2.14),\\
\text{TF}     &:\ \text{NLL} = 2.9170\ (\text{PPL} = 18.49).
\end{align*}

As at shorter horizons, latent autoregression yields trajectories more
compatible with the causal GP prior. As the latent horizon increases, the
correlation structure imposed by the prior penalizes the non-AR variant more
strongly, whose i.i.d.-like behavior becomes statistically incompatible. The AR
vs.\ non-AR gap, subtle at $L_{\text{gen}} = 1024$, re-emerges here in a
pronounced way.

\paragraph{Sequential structure.}
Surface metrics confirm this tendency:
\begin{align*}
\text{GP-AR:}\ &\text{rep2}=0.016,\quad \text{rep3}=0,\quad \text{consec}=0.111,\\
\text{GP-noAR:}\ &\text{rep2}=0,\quad \text{rep3}=0,\quad \text{consec}=0.
\end{align*}

The AR latent exhibits weak but genuine structure without degenerate loops,
whereas the non-AR variant remains quasi-stationary. By contrast, the
Transformer exhibits classical token-level repetition patterns.

\paragraph{Token-level AR metrics.}
GPT-2 assigns extremely high perplexities to both GP-VAE variants
($\sim 10^{5}$) while strongly favoring the Transformer. These values reflect
the non-autoregressive nature of the GP-VAE and confirm that standard
token-autoregressive metrics are ill-suited to evaluate a parallel latent-GP
model, especially at long horizons.

\paragraph{Summary.}
This experiment shows that latent autoregression remains robust and increasingly
exploits the GP prior as the latent path length grows. At $L_{\text{gen}}=2048$,
GP-VAE AR preserves internal coherence, whereas the non-AR variant becomes more
discordant with the structure imposed by the prior. The Transformer, for its
part, degrades in internal probabilistic structure, despite being favored by
external token-autoregressive metrics.

\subsection{AR vs.\ non-AR on \textsc{WikiText-103}
(\texorpdfstring{$L_{\text{gen}} = 3072$}{Lgen=3072})}

On \textsc{WikiText-103} with a very long horizon ($L_{\text{gen}}=3072$), the
latent-AR GP-VAE remains stable: validation internal perplexity is
$\text{PPL}\approx 3.83$ and continuation perplexity is $\text{PPL}\approx 3.96$,
with KL remaining at $12$ nats and no sequential collapse.

Intrinsic continuation metrics confirm the advantage of latent autoregression
over the non-AR variant, while the Transformer baseline becomes substantially
less coherent under its own distribution:
\begin{table}[htbp]
  \centering
  \small
  \begin{tabular}{lcc}
    \toprule
    Model & NLL & PPL \\
    \midrule
    GP-VAE AR     & $0.6840$ & $1.98$ \\
    GP-VAE non-AR & $0.7809$ & $2.18$ \\
    Transformer   & $3.6684$ & $39.19$ \\
    \bottomrule
  \end{tabular}
  \caption{Own-model continuation metrics on \textsc{WikiText-103}
  (\texorpdfstring{$L_{\text{gen}} = 3072$}{Lgen=3072}).}
  \label{tab:w103_ar_vs_nonar_3072}
\end{table}

The latent autoregressive variant therefore maintains significantly more
predictable trajectories than the i.i.d.\ latent, even at very long horizons,
whereas the Transformer degrades sharply in internal probabilistic structure.
As at other context scales, GPT-2 assigns massive perplexities to both GP-VAE
variants (on the order of $10^{5}$) while favoring the Transformer (PPL
$\approx 125$), confirming that token-autoregressive metrics are not a relevant
criterion for evaluating a parallel latent-GP model.


\section{Direct Comparison on \textsc{WikiText-2}: GP-VAE AR vs.\ Transformer}
\label{sec:w2-gpvs-tf}

We complement the previous analysis with a controlled experiment on
\textsc{WikiText-2}, in which we directly compare the GP-VAE with latent
autoregression to a token-level autoregressive Transformer.

The protocol is as follows. The \textsc{WikiText-2 raw} corpus is loaded via the
\texttt{datasets} API, tokenized using the GPT-2 tokenizer, and concatenated
into fixed-length blocks. The GP-VAE with latent autoregression uses exactly the
same TCN encoder, parallel decoder, and training procedure (ELBO objective,
KL cap at 8 nats, and $\beta$ annealing) as in the previous experiments.

\paragraph{Perplexity: methodological clarification.}
The perplexities reported for the GP-VAE correspond to \emph{conditional}
perplexities (see Section~\ref{subsec:metrics}) and do not arise from an
autoregressive factorization.
In contrast, the Transformer's perplexity corresponds to the standard
token-level quantity
\[
p(x_t \mid x_{<t}).
\]
These two quantities are therefore \emph{not directly comparable}.
Meaningful comparison instead relies on:
\begin{itemize}
    \item continuation-based metrics (NLL, PPL, coherence),
    \item qualitative behavior (repetition patterns, degeneracy),
    \item and the effect of latent autoregression.
\end{itemize}

On this corpus, the GP-VAE with latent autoregression achieves a validation
perplexity of
\[
\text{PPL(val)} \approx 2.25, \qquad \text{ELBO/token} \approx -8.27,
\]
while the Transformer baseline reaches a validation loss of approximately
$3.18$ nats per token, corresponding to
\[
\text{PPL(val)} \approx 24.0.
\]
The GP-VAE perplexity should be interpreted as an \emph{internal} coherence
measure rather than a directly comparable language-model likelihood.

Beyond aggregate scores, we evaluate both models using generation-based metrics.
For a set of prompts drawn from the test split, we generate continuations of
lengths $L \in \{32, 64, 128, 256, 512, 1024, 2048, 3072\}$ and measure:

\begin{itemize}
  \item repetition statistics (bigram/trigram repetition, fraction of repeated tokens),
  \item GPT-2–based perplexity and rarity metrics,
  \item qualitative properties of the generated sequences.
\end{itemize}

For the GP-VAE with latent autoregression, repetition remains moderate across
all sequence lengths: bigram repetition typically lies in the range
$0.06$--$0.30$, trigram repetition in $0.01$--$0.15$, and the fraction of exact
loops remains close to zero. No catastrophic degeneration is observed even at
$L = 3072$. Nevertheless, GPT-2 assigns high perplexities (often between
$10^{3}$ and $10^{5}$), reflecting the mismatch between the latent generative
process and the autoregressive scoring model.

By contrast, the Transformer baseline exhibits rapid collapse. Already at
$L=32$, repetition rates approach one, loop frequency reaches unity, and the
generated sequences degenerate into short repeating patterns. Paradoxically,
these degenerate outputs receive very low perplexity scores from GPT-2, since
they are highly predictable under an autoregressive model.

These results highlight a central point: standard token-level perplexity is not
a reliable indicator of generative quality when comparing models with
fundamentally different factorization structures. The GP-VAE with latent
autoregression produces coherent, non-collapsing sequences whose structure is
incompatible with autoregressive evaluation, while the Transformer optimizes
precisely for such evaluation even at the cost of severe mode collapse.

In summary, the experiments on \textsc{WikiText-2} demonstrate that latent
autoregression yields stable long-range generation and avoids the degeneracies
observed in token-autoregressive models, at the cost of producing samples that
are penalized by standard autoregressive metrics.


\section{Central Analysis: AR vs.\ Non-AR in the Latent Space}
\label{sec:latent-ar-vs-nonar}

We summarize here the three internal diagnostics that highlight the critical
role played by autoregressive dynamics in the latent space.

\paragraph{(A) Compatibility with the GP prior.}
The average log-density of latent trajectories evaluated under the correlated
Gaussian prior is substantially higher for the latent-autoregressive variant
($\log p_{\text{GP}}(z) \approx 2.3 \times 10^{3}$), indicating strong
compatibility with the causal Gaussian process. In contrast, the non-AR variant
exhibits a very low log-density
($\log p_{\text{GP}}(z) \approx -2.7 \times 10^{7}$), reflecting a severe mismatch
with the correlated prior.

The resulting gap in log-density between the two regimes, on the order of
$2.7 \times 10^{7}$ nats, demonstrates that trajectories produced without
latent autoregression are fundamentally incompatible with the geometry imposed
by the GP prior.

\paragraph{(B) Temporal structure of the latent trajectories.}
The mean cosine auto-correlation as a function of lag shows that the AR variant
exhibits strong short-range correlations (lag~1: $\rho \approx 0.57$), which
gradually decay with increasing lag—an indicator of coherent temporal dynamics.
In contrast, the non-AR variant remains near zero correlation at all lags
($\rho \approx 0.00$), consistent with i.i.d.\ noise.

\paragraph{(C) Magnitude of latent variations.}
The average step size $\|z_t - z_{t-1}\|$ further highlights this difference.
The AR model produces smoother trajectories
($4.87 \pm 0.98$), whereas the non-AR variant exhibits significantly larger
step sizes ($8.46 \pm 0.74$), characteristic of unstructured noise.

\paragraph{Summary.}
Taken together, these diagnostics show that latent autoregression does not
merely provide a marginal regularization effect: it fundamentally shapes the
temporal structure of the latent space. The AR model yields coherent, smooth
latent trajectories consistent with the Gaussian-process prior, whereas the
non-AR variant produces incoherent dynamics that violate the assumptions of the
underlying prior.


\section{Methodological Remark: Structural Bias Induced by GPT-2 as an External Judge}
\label{sec:bias-gpt2}

It is important to emphasize that the evaluation of linguistic plausibility in our study relies on an external model—GPT-2—whose probabilistic factorization is strictly autoregressive. While this choice is methodologically reasonable, as it provides a common reference for comparing models with different internal structures, it also introduces a structural bias that must be made explicit.

Indeed, GPT-2 estimates the probability of a sequence according to a factorization of the form
\[
p(x_t \mid x_{<t}),
\]
learned during autoregressive pretraining. Sequences that conform to the local statistical regularities of such models—namely, token-by-token dependencies and short-range conditional structure—will therefore be assigned higher likelihoods. In contrast, sequences generated by non-autoregressive models, such as the GP-VAE studied here, do not follow this factorization and are not constrained by the same inductive biases. Their dynamics are governed by global latent variables rather than local token-level prediction.

As a consequence, the use of GPT-2 as an external evaluator introduces a \emph{compatibility bias}: an autoregressive model is favored not only because it may produce linguistically plausible text, but also because its internal generative structure aligns with that of the evaluator. Conversely, a non-autoregressive generative model may receive a low likelihood under GPT-2 even when its outputs are coherent, simply because its generative mechanism does not match the autoregressive assumption.

This bias does not invalidate the evaluation, but it constrains its interpretation. GPT-2–based perplexity should not be viewed as an absolute measure of linguistic quality, but rather as an indicator of statistical proximity to the family of autoregressive language models. In this work, we therefore use GPT-2–based metrics primarily to compare relative behaviors (e.g., AR vs.\ non-AR variants) rather than to claim absolute superiority in linguistic modeling.

We deliberately use GPT-2 scores as \emph{relative} indicators within a given model family (e.g., GP-AR vs.\ GP-noAR), while grounding our main conclusions in intrinsic continuation metrics, latent-space structure, and long-horizon stability—quantities that more faithfully reflect the modeling assumptions and objectives of our approach.


\section{Overall Synthesis of Results}
\label{sec:synthesis}

The experiments conducted on \textsc{WikiText-2} and \textsc{WikiText-103} reveal a
clear contrast between autoregressive and non-autoregressive latent dynamics
within the same GP-VAE architecture.

When the latent process is governed by a causal Gaussian process, the model
learns temporally correlated trajectories that are compatible with the prior
and maintains stable behavior over long generation horizons. In contrast, the
non-autoregressive ablation yields latent representations that behave like
independent noise, are incompatible with the GP prior, and lead to systematic
degradation in long-range behavior.

These findings are consistently reproduced across two independent
implementations of the model and across datasets of different scales. They
indicate that, in the studied regime, latent dynamics play a central role in
ensuring sequential coherence, independently of the expressive power of the
decoder.


\section{Discussion}

It is important to emphasize that the conclusions of this work do not rely on a
specific implementation of the GP-VAE, but rather on consistent phenomena
observed across two independent implementations, each used under the constraints
of its respective experimental setup.

A central finding of this study is that autoregression in the latent space does
not act merely as an auxiliary regularizer, but instead plays a structurally
defining role in the model’s behavior. When this latent dynamics is removed, the
GP-VAE retains the ability to perform local reconstruction, yet loses global
temporal coherence, revealing a clear dissociation between local fidelity and
long-range stability.

This observation is conceptually significant: it shows that a substantial part
of sequential structure can be transferred from the symbolic token space to a
continuous latent space, without relying on explicit autoregressive decoding.
In this setting, the Gaussian prior does more than regularize latent magnitudes:
it imposes a temporal geometry that actively shapes learning dynamics and
stabilizes generation over long horizons.

The contrast observed between internal coherence and external evaluation further
highlights an important methodological limitation. A model may be penalized by an
autoregressive evaluator not because it lacks linguistic coherence, but because
its generative factorization fundamentally differs from that of the evaluator.
This discrepancy underscores the need for evaluation criteria that are better
aligned with the structural assumptions of non-autoregressive or latent-variable
models.

Overall, these results suggest that the expressive power of a language model is
not solely determined by the depth or size of its decoder, but also by the nature
of the probabilistic structure imposed on its latent space. This perspective
opens the door to hybrid architectures in which sequential coherence is carried
primarily by latent dynamics, while the decoder remains lightweight and
parallelizable.

\section{Conclusion and Perspectives}
\label{sec:conclusion}

This work provides an empirical feasibility study of purely latent autoregressive
models for language modeling. Under controlled architectural and capacity
conditions, our experiments demonstrate that the dynamics imposed in the latent
space play a decisive role in shaping the sequential behavior of the model.

When the latent process follows a causal Gaussian prior, the GP-VAE learns temporally correlated trajectories that are consistent with the prior and remains stable over the longest horizons evaluated in this work (up to $L_{\text{gen}} = 3072$). In contrast, the non-autoregressive ablation—despite being trained under the same variational objective—produces latent representations akin to independent noise and leads to a systematic degradation of long-range coherence.

These differences in latent structure translate directly into the properties of
the generated text. The latent autoregressive regime preserves diversity while
avoiding catastrophic looping, whereas the non-AR variant collapses as the
generation horizon increases. The comparison with an autoregressive Transformer
further highlights the limitations of standard perplexity-based metrics, which may
favor locally predictable but globally degenerate behaviors.

Taken together, these results indicate that a significant portion of sequential
structure can be carried by a correlated latent process, independently of a
powerful autoregressive decoder. In this sense, autoregressivity in the latent
space is not a secondary design choice but a central mechanism governing
long-range coherence and stability.

\paragraph{Outlook.}
Future work may explore richer latent dynamics, longer context windows, and hybrid
architectures that combine latent correlation with lightweight autoregressive
decoding. Extending the evaluation to broader benchmarks and human-centered
criteria will also be essential to better assess the strengths and limitations of
this class of models.

\end{document}